\documentclass[journal,comsoc]{IEEEtran}
\usepackage[T1]{fontenc}
\usepackage{cite}

\ifCLASSINFOpdf
   \usepackage[pdftex]{graphicx}
\else
   \usepackage[dvips]{graphicx}
\fi

\usepackage{bm}
\usepackage{amsmath}
\usepackage[cmintegrals]{newtxmath}

\usepackage[numbers]{natbib}
\usepackage{booktabs}
\usepackage{multirow}
\usepackage{multicol}
\usepackage{amsmath}

\usepackage{algorithm}
\usepackage{algorithmicx}
\usepackage{algpseudocode} 

\usepackage{mathtools}
\usepackage{hyperref}

\mathchardef\mhyphen="2D

\begin{document}
%
\title{Cross-Question Method Reuse in Large Language Models: From Word-Level Prediction to Rational Logical-Layer Reasoning}

\author{Hong~Su
\IEEEcompsocitemizethanks{\IEEEcompsocthanksitem H. Su is with the School of Computer Science, Chengdu University of Information Technology, Chengdu, China.\\
 E-mail: suguest@126.com. \\
\protect\\
}
\thanks{}}

\markboth{Journal of \LaTeX\ Class Files,~Vol.~14, No.~8, August~2015}%
{Shell \MakeLowercase{\textit{et al.}}: Bare Demo of IEEEtran.cls for IEEE Communications Society Journals}
%

\maketitle

\begin{abstract}
Large language models (LLMs) have been widely applied to assist in finding solutions for diverse questions. Prior work has proposed representing a method as a pair of a question and its corresponding solution, enabling method reuse. However, existing approaches typically require the questions to be highly similar. In this paper, we extend the scope of method reuse to address questions with low similarity or with hidden similarities that are not explicitly observable.
For questions that are similar in a general–specific sense (i.e., broader or narrower in scope), we propose to first separate the question and solution, rather than directly feeding the pair to the LLM. The LLM is then guided to adapt the solution to new but related questions, allowing it to focus on solution transfer rather than question recognition. Furthermore, we extend this approach to cases where questions only share partial features or hidden characteristics. This enables cross-question method reuse beyond conventional similarity constraints.
Experimental verification shows that our scope-extension approach increases the probability of filtering out reusable solutions, thereby improving the effectiveness of cross-question method reuse.
\end{abstract}


\IEEEpeerreviewmaketitle

\section{Introduction}
Large language models (LLMs) \cite{naveed2025comprehensive} are increasingly used to assist with complex tasks such as writing, reasoning \cite{plaat2024reasoning}, and problem solving \cite{renze2024benefits}. Traditional LLMs are primarily trained at the word level to predict the next token given a context, or through masked word prediction by filling in missing tokens \cite{kumar2024large}. While effective, this word-level training largely reflects statistical co-occurrence rather than higher-level logical reasoning, resembling intuition or pattern-matching more than rational decision making.

In contrast, methods operate at a higher logical level and can be represented as structured procedures for solving questions. For example, different methods may be compared or evaluated based on their effectiveness. However, current transformer-based LLMs often struggle with method-level reasoning. If a widely used method appears frequently in training data, the model may prefer it even when it is not optimal. To address this, Su et al.\ \cite{su2025method} proposed a method-based solution framework, in which each method is represented as a pair consisting of a question and its corresponding solution.

An important advantage of method-based reasoning is that methods can be reused across different questions. This enables the generation of creative solutions, similar to how a human might apply knowledge from one problem to address a new, seemingly unrelated one. We refer to this process as \textit{cross-question method reuse}. Unlike word-level reasoning, cross-question reuse enables LLMs to rationally transfer solutions even when no explicit training patterns exist for the target problem.

To illustrate, consider the following example where a solution is borrowed from another method that is not similar at the word level. Tom wishes to reset the usage time of his hard disk by himself, without sending it to others, since he does not want anyone else to access its contents. However, he does not possess the necessary specialized tools. At first glance, no similar case seems available. Yet Tom recalls a past experience (\textit{expMP3}): his friend Mary needed a slower version of an MP3 file, but no direct download was available. Even after paying on the official website, only a proprietary format was provided. Eventually, Tom resolved the problem by purchasing the correct file from a goods-exchange website. By analogy, he later addressed his disk-reset issue in a similar way—purchasing the appropriate reset tool online. This example demonstrates how a method from one domain can be “borrowed” to solve a seemingly unrelated problem, despite minimal lexical similarity.

\begin{figure}[htb]
    \centering
    \includegraphics[width=3.5in]{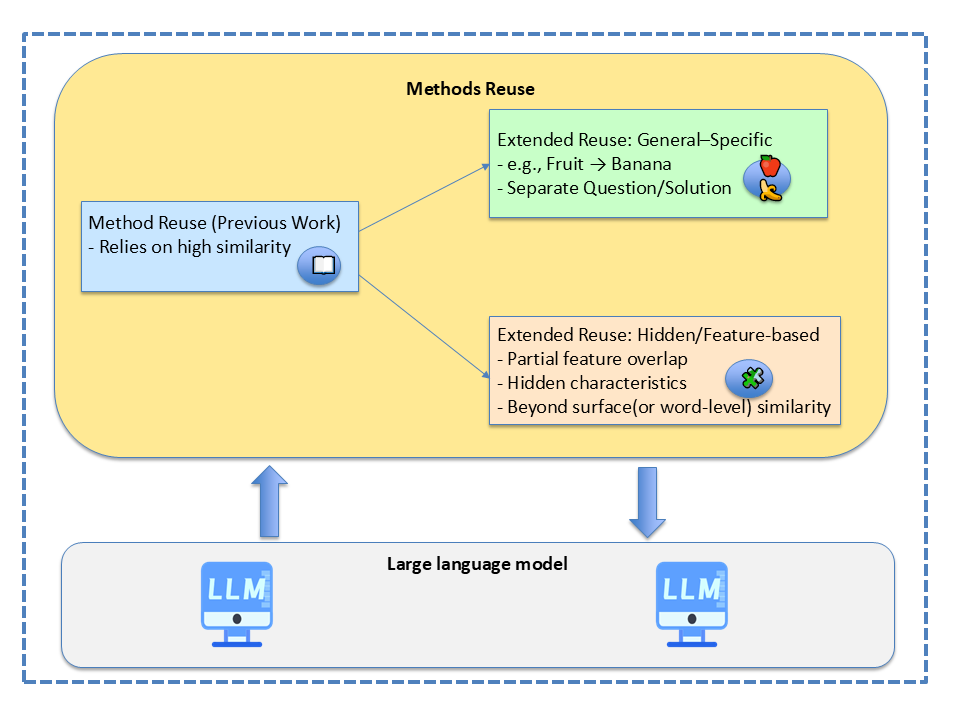}
    \caption{Extended scope of method reuse: from high-similarity cases to general–specific mappings and hidden/feature-based relationships.}
    \label{fig_intro_explain}
\end{figure}

Building on the method-based LLM framework proposed in \cite{su2025method}, which focuses on reusing methods across highly similar questions, this paper extends the scope of reuse to include less explicit or hidden similarities. In particular, we consider cases where methods are not aligned at the word level. For questions that exhibit general–specific relationships (e.g., fruit vs.\ banana), we propose separating the question and solution components prior to prompting, thereby guiding the model to reason about how a solution can be transferred to a related but more specific problem. Beyond such cases, we further extend reuse to scenarios without explicit hierarchical relations, but where questions \textbf{share partial features or latent characteristics}. This extension is illustrated in Figure~\ref{fig_intro_explain}, which highlights how the proposed approach broadens the applicability of method reuse beyond direct or surface-level similarity.

The main contributions of this paper are as follows:
1) We extend method reuse to a broader scope by enabling cross-question reuse when questions are related through general–specific or parallel similarities.
2) We introduce feature-based reuse, which supports method transfer when questions share only partial or hidden similarities.
3) We propose the concept of a \textit{method of method} (MoM), where higher-level methods are automatically retrieved to validate, refine, or improve the effectiveness of currently applied methods.

The remainder of this paper is organized as follows. Section II reviews related work. Section III introduces the proposed Cross-Question Method Reuse model. Section IV presents experimental verification and analysis. Finally, Section V concludes with a summary.

\section{Related Work} \label{sec_related_work}
The proposed framework for cross-question method reuse relates to several active areas of research, including reasoning with large language models (LLMs), method and knowledge reuse, transfer learning, and meta-reasoning with self-improving strategies.

\subsection{Reasoning with Large Language Models}
Recent studies have demonstrated that LLMs benefit from explicit reasoning structures. Chain-of-thought prompting \cite{wei2022chain} \cite{wang2024chain} guides models to generate intermediate steps, improving performance on arithmetic and logical reasoning tasks. Self-consistency \cite{wang2022self} further enhances robustness by sampling multiple reasoning paths and aggregating the results. Extensions such as tree-of-thoughts \cite{yao2023tree} \cite{lin2023solving} and graph-of-thoughts \cite{besta2024graph} generalize this idea to more complex search structures.

While these approaches improve reasoning fidelity, they typically operate within a single problem space and do not emphasize systematic reuse of methods across different questions. Our work complements this line of research by introducing a structured approach for transferring solutions from one question to another, even when similarity is weak or implicit.

\subsection{Method Representation and Reuse}
The idea of representing knowledge as reusable components has been explored in both symbolic AI and LLM-based approaches. Prior work has suggested decomposing solutions into question–solution pairs \cite{su2025method}, allowing methods to be reused across different questions and enabling automatic extraction of reusable strategies, rather than relying solely on pre-defined reasoning steps such as chain-of-thought prompting.

Similar principles also appear in program synthesis and automated planning, where abstract operators or subroutines are reused to solve new tasks \cite{prountzos2015synthesizing} \cite{naciri2015multi}. However, these approaches typically assume explicit similarity between tasks. In contrast, our contribution extends this paradigm by supporting reuse across less obvious connections, introducing both relationship-based reuse (general–specific or parallel similarity) and feature-based reuse (partial or hidden similarity).

\subsection{Transfer Learning and Meta-Reasoning}
Transfer learning has been widely studied as a mechanism for adapting models across domains or tasks. Pretrained models such as T5 \cite{li2024table} and GPT-4 \cite{yenduri2024gpt} demonstrated that fine-tuning or in-context learning \cite{thomas2024retrieval} can enable rapid task transfer, though often at the cost of large-scale data or explicit demonstrations. Case-based reasoning (CBR) \cite{schultheis2023overview} provides a complementary approach, solving new problems through analogy with prior examples. In the LLM era, CBR has been revisited through retrieval-augmented prompting \cite{dong2025understand}, where examples are retrieved and reused to guide few-shot inference. While these approaches assume the existence of strongly similar prior cases, our framework extends reuse to situations where connections are weak, partial, or hidden, thereby broadening the range of applicable prior knowledge.

A further line of related work focuses on meta-reasoning, where LLMs not only solve tasks but also reflect on or evaluate their own reasoning. Reflexion \cite{shinn2023reflexion} enables iterative self-feedback for improvement, while debate-style frameworks \cite{du2023improving} employ multiple models or agents to critique and refine outputs. More recently, self-verification and double-checking strategies \cite{diao2023active} have shown that LLMs can serve as their own evaluators. Our concept of methods-of-methods is aligned with this direction: higher-depth methods operate on other methods to validate, regulate, or enhance them, thereby improving robustness and reliability.

\subsection{Summary}
In summary, prior research has advanced LLM reasoning fidelity, explored method decomposition, leveraged case reuse, and developed meta-reasoning strategies. However, most of these approaches rely on strong or explicit similarities between tasks. Our work differs by broadening the scope of reuse to include cross-question scenarios where connections are implicit, feature-based, or dependency-driven. By structuring problems as question–solution pairs and enabling both relationship- and feature-based reuse, the proposed framework offers a novel pathway for enhancing the adaptability and versatility of LLMs in solving diverse and previously unseen problems.

\section{Model for Cross-Question Method Reuse} \label{sec_think_model}
In the method-based LLM model, a method is defined as a pair consisting of a question and its corresponding solution \cite{su2025method}. This formulation aims to improve reusability, but in that prior work, the questions were required to be highly similar. In this paper, we extend the reuse scope to allow a method to be applied across a wider range of questions, including those that do not have an obvious relationship.

The motivation is that the method-based approach fits more scenarios when questions are the same or highly similar. However, if the questions are less or even not similar (suppose $Q_a$ and $Q_b$), it may still be possible to use the solution of $Q_a$ to address $Q_b$ when no direct solution for $Q_b$ exists. We define this kind of reuse as \textbf{cross-question method reuse}, or equivalently, a \textbf{method association}. The idea is analogous to how a person, when facing a new unsolved problem, may attempt to adapt solutions from earlier, seemingly unrelated problems.

We propose two categories of reuse strategies: the \textit{method relationship way} and the \textit{feature way}.

\subsection{Method Relationships} \label{sec_relationship}
Different types of relationships can exist among methods with respect to their sets of solvable questions. In this section, we focus on two primary categories: (1) \textit{general and specific methods}, where the scope of one method is contained within another, and (2) \textit{parallel methods}, where methods address different subsets of the same broader category.

Let two methods $M_a$ and $M_b$ solve the sets of questions $Q_{ma}$ and $Q_{mb}$, producing solutions $S_{ma}$ and $S_{mb}$, respectively.

\subsubsection{General and Specific Methods}
Some methods have a broader scope of applicability and can solve more questions. We call these methods \textit{general}. If $M_a$ and $M_b$ satisfy the relationship in \eqref{eq_more_general}, then $M_a$ is more general than $M_b$, and $M_b$ is more specific than $M_a$. Reusing a more general method to solve the question of a more specific method is referred to as \textit{vertical reuse}.

\begin{equation}\label{eq_more_general}
    Q_{ma} \supset Q_{mb}
\end{equation}

For example, to determine whether a plantain is fresh, one can reuse the method for judging the freshness of bananas, since the banana method has broader coverage.

\subsubsection{Parallel Methods}
Two methods are \textit{parallel} if their question sets are disjoint subsets of the same broader category. Formally, there exists a superset $Q_g$ such that
\begin{equation}\label{eq_parallel}
    Q_{ma} \cap Q_{mb} = \varnothing, \quad Q_{ma} \subset Q_g, \quad Q_{mb} \subset Q_g.
\end{equation}
In this case, applying $S_{ma}$ to solve $Q_{mb}$ (or vice versa) constitutes \textit{parallel method reuse}, which is a form of \textit{horizontal reuse}.

For example, the questions “How to judge whether a banana is fresh?” and “How to judge whether an apple is fresh?” are parallel. Both belong to the broader category of “fruit freshness,” but neither subsumes the other. Reusing a solution across such parallel cases illustrates horizontal transfer of methods.

In another situation, if a single solution $S$ can solve two questions $Q_1$ and $Q_2$ without obvious shared characteristics, $Q_1$ and $Q_2$ may still be regarded as parallel with respect to $S$. However, this work does not emphasize such cases, which were considered in prior studies \cite{su2025method}, since our primary objective is to identify solutions for previously unknown questions.

\subsection{Feature-Based Reuse} \label{sec_feature}
In the relationship-based approach, the questions (or solutions) of methods must exhibit an explicit logical connection (vertical or horizontal). By contrast, \emph{feature-based reuse} applies when only certain aspects of two questions are shared, even if the questions themselves are not directly similar. If these shared aspects allow a solution from one question to be applied to another, we refer to them as \textbf{features}.

\paragraph*{Feature space and extraction}
Let $\mathcal{F}$ denote a (possibly hybrid) feature space comprising measurable attributes (e.g., temperature, height) and descriptive attributes (captured via textual embeddings). For a question $Q$, we define its features as
\begin{equation}
F(Q) \subseteq \mathcal{F}, \qquad F(Q) = F_{\text{meas}}(Q) \cup F_{\text{text}}(Q),
\end{equation}
where $F_{\text{meas}}(Q)$ are explicit numeric attributes and $F_{\text{text}}(Q)$ are features derived from a learned encoder $h(\cdot)$ over text (e.g., an LLM):
\begin{equation}
F_{\text{text}}(Q) = \{\, f \in \mathcal{F} \mid f \leftrightarrow h(Q) \,\}.
\end{equation}

\paragraph*{Feature similarity and reuse condition}
Given two questions $Q_a$ and $Q_b$, their feature similarity is defined as
\begin{equation}
\mathrm{Sim}_{\text{feat}}(Q_a,Q_b) = S\!\big(F(Q_a),\,F(Q_b)\big),
\end{equation}
where $S(\cdot,\cdot)$ may be the Jaccard index for symbolic features, cosine similarity for embeddings, or a weighted hybrid.

A solution $S_a$ associated with $Q_a$ is reused for $Q_b$ if
\begin{equation}
\mathrm{Reuse}_{\text{feat}}(Q_b; S_a) =
\begin{cases}
1, & \text{if } \parbox[t]{0.6\linewidth}{\raggedright$\mathrm{Sim}_{\text{feat}}(Q_a,Q_b) \ge \tau$ and  $\mathrm{Valid}(S_a, Q_b)=1$}, \\
\\
0, & \text{otherwise},
\end{cases}
\label{eq:feature-reuse}
\end{equation}
where $\tau \in [0,1]$ is a similarity threshold and $\mathrm{Valid}(S_a,Q_b)$ is a predicate that checks whether $S_a$ is logically valid for $Q_b$ (e.g., domain constraints, rules, or verification by another LLM).

This condition distinguishes feature-based reuse from relationship-based reuse: instead of requiring set inclusion among question scopes, reuse is triggered when the degree of feature overlap is sufficiently high.

\subsubsection{Partial Feature Matching}
If the question of one method $Q_s$ does not fully match a target question $Q_t$, then $Q_t$ is said to \emph{partially match} the features of $Q_s$. In such cases, reuse is guided by the condition in Eq.~\eqref{eq:feature-reuse}: a solution $S_s$ associated with $Q_s$ is reused for $Q_t$ if their feature similarity exceeds a threshold and the solution is validated for the new context.

This ensures that reuse is not based solely on surface overlap, but also on whether the candidate solution can be plausibly applied to the target question.

When no direct solution exists for $Q_t$, the method with the highest feature similarity may be applied as a candidate. If this attempt fails (i.e., the validity check in Eq.~\eqref{eq:feature-reuse} does not hold), other candidates are considered sequentially.

To handle large feature spaces efficiently, two strategies are adopted:
\begin{itemize}
    \item \textbf{Global maximum:} select the method with the highest similarity score overall and attempt reuse once, as prescribed by Eq.~\eqref{eq:feature-reuse}.
    \item \textbf{Relative maximum:} select all methods above a similarity threshold $\tau$ and attempt them in descending order of similarity; if none succeed, the threshold is gradually relaxed until either reuse succeeds or the attempt budget $K$ is exhausted.
\end{itemize}

\begin{algorithm}[t]
\caption{Partial Feature Matching for Cross-Question Reuse}
\label{alg:pfm}
\begin{algorithmic}[1]
\Require Target question $Q_t$; method set $\mathcal{M}=\{M_i=(Q_i,S_i)\}$ with feature representations; similarity function $\mathrm{sim}(\cdot,\cdot)$; mode $\in\{\textsc{Global},\textsc{Relative}\}$; threshold $\tau$; relaxation step $\Delta\tau$; minimum threshold $\tau_{\min}$; attempt budget $K$
\Ensure A reused method $M^\star$ or \textsc{None}
\State Compute feature vector $f_t \leftarrow \mathrm{Feat}(Q_t)$
\For{each $M_i \in \mathcal{M}$}
  \State $f_i \leftarrow \mathrm{Feat}(Q_i)$; \quad $s_i \leftarrow \mathrm{sim}(f_t,f_i)$
\EndFor
\If{mode = \textsc{Global}}
  \State $i^\star \leftarrow \arg\max_i s_i$
  \State \Return $M_{i^\star}$ if $\textsc{Apply}(S_{i^\star},Q_t)=\textsc{Success}$ else \textsc{None}
\Else \Comment{\textsc{Relative} mode}
  \State $A \leftarrow$ indices sorted by $s_i$ in descending order
  \State $k \leftarrow 0$
  \While{$k < K$}
    \State $C \leftarrow \{i \in A \mid s_i \ge \tau\}$
    \For{each $i \in C$}
      \If{$\textsc{Apply}(S_i,Q_t)=\textsc{Success}$} \Return $M_i$ \EndIf
      \State $k \leftarrow k + 1$; \quad \If{$k \ge K$} \Return \textsc{None} \EndIf
    \EndFor
    \State $\tau \leftarrow \max(\tau - \Delta\tau,\ \tau_{\min})$ \Comment{Relax threshold}
    \If{$C=\varnothing$ and $\tau=\tau_{\min}$} \Return \textsc{None} \EndIf
  \EndWhile
\EndIf
\end{algorithmic}
\end{algorithm}

Algorithm~\ref{alg:pfm} operationalizes Eq.~\eqref{eq:feature-reuse}. In \textsc{Global} mode, the candidate with the maximum similarity is directly tested for reuse. In \textsc{Relative} mode, multiple candidates are evaluated in sequence, with progressively relaxed thresholds, ensuring that reuse remains both accurate and computationally efficient.

\subsubsection{Hidden Characteristics and Emerging Features}
Not all similarities can be captured through explicit features. Some are \textbf{hidden characteristics}—semantic or contextual overlaps embedded in descriptive text. For example, the questions “Why is the website slow?” and “Why is the system response delayed?” may share hidden performance-related characteristics, even though no explicit numeric feature links them.

In these cases, reuse is guided by a two-step process:
\begin{enumerate}
    \item Use an LLM to assess whether the target question $Q_t$ and a candidate $Q_s$ exhibit latent semantic similarity.
    \item If so, use the LLM to evaluate whether $S_s$ can plausibly be applied to $Q_t$.
\end{enumerate}

If both conditions hold, $S_s$ is reused for $Q_t$.

When no explicit or hidden features can be identified, especially for new or emerging concepts, the entire method $M=(Q,S)$ may itself be stored as a \textbf{template feature}. Define a meta-feature embedding
\begin{equation}
\phi(M) = g(Q,S),
\end{equation}
where $g(\cdot)$ jointly encodes the question and solution. For a new question $Q_t$, similarity is measured as
\begin{equation}
\mathrm{Sim}_{\text{meta}}(M,Q_t) = \cos\!\big(\phi(M),\,h(Q_t)\big),
\end{equation}
and reuse is permitted if
\begin{equation}
\mathrm{Sim}_{\text{meta}}(M,Q_t) \ge \tau_{\text{meta}} \quad \text{and} \quad \mathrm{Valid}(S,Q_t)=1.
\end{equation}

With feature-based reuse, methods can be applied across questions in the following cases:
\begin{itemize}
    \item \textbf{Partial feature matching:} reuse enabled by explicit, measurable overlap between features.
    \item \textbf{Hidden characteristics:} reuse guided by semantic or contextual similarities that are inferred through LLM-based reasoning rather than direct measurements.
    \item \textbf{Emerging features (entire-method-as-feature):} reuse applied when neither explicit nor hidden features are available, by treating the entire method $(Q,S)$ as a reusable template for future cases.
\end{itemize}

\subsection{Global Methods}
Not all methods are designed for a single, specific question. Some exhibit broad applicability across many situations—these are termed \textit{global methods}. A global method captures general behavioral or procedural rules that can support other, more specific methods. For example, the principle “arrive 30 minutes early for important events” is a global method that can assist different tasks such as attending meetings, teaching classes, or catching a flight.  

\paragraph*{Definition and Representation}
Let $\mathcal{M}_g = \{G_1, G_2, \ldots, G_n\}$ denote the set of all global methods.  
Each global method $G_i$ can be represented as a pair
\begin{equation}
G_i = (Q_{g_i}, S_{g_i}),
\end{equation}
where $Q_{g_i}$ is the general question or objective (e.g., “how to ensure punctuality”), and $S_{g_i}$ is the corresponding general solution (e.g., “arrive 30 minutes early”).  

For any specific method $M_j = (Q_j, S_j)$, we define the applicability of a global method $G_i$ as:
\begin{equation}
\mathrm{Apply}(G_i, M_j) = 
\begin{cases}
1, & \text{if } \mathrm{Valid}(S_{g_i}, Q_j) = 1,\\
0, & \text{otherwise.}
\end{cases}
\end{equation}
If $\mathrm{Apply}(G_i, M_j)=1$, then $G_i$ can be integrated into $M_j$ as an auxiliary step or heuristic.

\paragraph*{Integration and Maintenance}
When a method $M_j$ is executed, the system first checks whether any global method $G_i \in \mathcal{M}_g$ satisfies the reuse condition in Eq.~(8). If so, $G_i$ is incorporated into $M_j$ as an additional operational step:
\begin{equation}
S'_j = S_j \oplus S_{g_i},
\end{equation}
where $\oplus$ denotes composition or augmentation of the procedural steps.  
Once a global method has been successfully integrated and validated, it becomes a permanent subcomponent of $M_j$, eliminating the need to check the global method list for subsequent executions. Only when new global methods are introduced is the list re-evaluated.

\paragraph*{Example}
Suppose $M_j$ corresponds to the method “plan travel to a conference.” A global method $G_i$ such as “prepare and depart 30 minutes early” is tested for applicability. If valid, $S'_j$ becomes a composite method that incorporates $G_i$ as a precautionary substep.  

Thus, global methods act as general-purpose procedural augmentations that increase the reliability, consistency, and rationality of method execution across diverse question domains.

\subsection{Method of Methods}
Not all methods directly solve a question. Those that directly solve a question are referred to as \textit{direct methods}. In contrast, some methods operate on other methods in order to facilitate or improve problem solving; these are called \textit{methods of methods} (MoM). Such methods provide measurement, validation, enhancement, or supplementary support for other methods rather than producing direct answers themselves.

For instance, when evaluating a complex mathematical formula, a direct method would compute the result directly:
\[
M_{\text{direct}} : Q \mapsto S,
\]
where $Q$ is the question and $S$ is the computed solution.
A method of methods, by contrast, might require performing the calculation in two independent ways and comparing the outcomes:
\[
M_{\text{MoM}} : M_{\text{direct}} \mapsto \mathrm{Verify}(M_{\text{direct}}(Q)).
\]

\paragraph*{Hierarchical organization}
Methods of methods can also be organized hierarchically by depth. Let $M^{(0)}$ denote a \textit{0-depth method} (direct method), mapping questions to solutions:
\[
M^{(0)} : Q \mapsto S.
\]
A \textit{1-depth method} $M^{(1)}$ operates on $M^{(0)}$:
\[
M^{(1)} : M^{(0)} \mapsto \widetilde{M}^{(0)},
\]
where $\widetilde{M}^{(0)}$ is the validated or refined version of $M^{(0)}$.
More generally, an $(i+1)$-depth method operates on an $i$-depth method:
\[
M^{(i+1)} : M^{(i)} \mapsto \widetilde{M}^{(i)}.
\]
If $M^{(i)}$ fails to provide a satisfactory solution, $M^{(i+1)}$ can be invoked to evaluate, refine, or regulate it. Thus, depth provides a principled way to escalate from direct problem solving toward increasingly meta-level operations.

Like direct methods, methods of methods also have associated questions, features, and solutions. Take
Double-calculation method as an example: Question = \emph{“How can we verify the correctness of a complex mathematical expression?”}; Solution = \emph{“Perform an independent secondary calculation and compare the results.”}

The discovery of a suitable method of methods is also question-driven. The search begins with direct methods identified using the approaches in Sections~\ref{sec_relationship} and~\ref{sec_feature}. If the direct method fails, yields inconsistent results, or stricter validation is requested, a method of methods is sought.

Formally, let $\mathcal{M}^{(0)}$ be the set of direct methods. For a question $Q$, if $M^{(0)}(Q)$ fails, the system escalates to depth $i+1$:
\[
\exists M^{(i)} \in \mathcal{M}^{(i)} \quad \text{such that} \quad M^{(i)}(M^{(i-1)}) \mapsto \widetilde{M}^{(i-1)}.
\]
Although both direct methods and methods of methods may process the same input $Q$, their objectives differ: direct methods aim to solve the problem itself, while methods of methods focus on evaluating, validating, or improving the methods being applied.



		
	
		

\section{Theoretical Justification of Cross-Question Method Reuse}
The central argument of this paper is that method reuse enables LLMs to operate at a higher logical layer, beyond the statistical word-level predictions that dominate current architectures. While traditional LLMs rely on token co-occurrence patterns, our framework structures knowledge as question–solution pairs, allowing solutions to be transferred across questions even when no direct surface-level similarity exists. This section presents a theoretical justification for the proposed approach, focusing on two key points: (1) method reuse functions at the logical layer above the LLM’s word-level processing, and (2) such reuse ensures rational, rather than purely statistical, application of solutions.

\subsection{Logical Layer Reuse}
Traditional LLMs are trained to minimize prediction error at the token level. Formally, given a sequence of tokens $w_1, w_2, \dots, w_t$, the model learns a distribution
\begin{equation}
    P(w_{t+1} \mid w_1, w_2, \dots, w_t),
\end{equation}
which captures statistical co-occurrence patterns. This objective is effective for language generation but does not directly encode higher-level reasoning.

By contrast, in our framework, a method is represented as a question–solution pair,
\begin{equation}
    M = (Q, S),
\end{equation}
where $Q$ denotes the question and $S$ the corresponding solution. Reuse across questions is then defined as finding a mapping
\begin{equation}
    R: (Q_a, S_a) \rightarrow (Q_b, S_a),
\end{equation}
such that the solution $S_a$ of method $M_a$ is applicable to a different question $Q_b$.

This introduces a logical abstraction above the token level. Instead of predicting $w_{t+1}$ directly, the model reasons over reusable units $(Q, S)$. For instance, the solution $S=\text{``judge freshness by picking time''}$, originally paired with $Q=\text{``How to judge whether a fruit is fresh?''}$, can be mapped to a new question $Q'=\text{``How to judge whether a banana is fresh?''}$. At the word level, “fruit” and “banana” overlap statistically; at the method level, the reuse is justified logically because $Q'$ is a specialization of $Q$.

Thus, method reuse operates at the logical layer:
\begin{equation}
    \text{Logical Reuse: } Q_b \in \text{Scope}(Q_a) \implies (Q_b, S_a) \in \mathcal{M},
\end{equation}
where $\mathcal{M}$ is the set of valid reusable methods. This formalization highlights that cross-question method reuse depends on logical relationships between questions, not only on token-level similarity.

\subsection{Rational Reuse}
In addition to operating at a logical layer, the proposed framework ensures that methods are reused \textit{rationally}, rather than merely according to statistical frequency. Traditional LLMs select continuations by maximizing token-level probability:
\begin{equation}
    \hat{w}_{t+1} = \arg\max_{w} P(w \mid w_1, w_2, \dots, w_t).
\end{equation}
This approach favors highly probable patterns in the training data, even when they are not the most effective solutions for a given question. For example, if a suboptimal method is more frequent in the corpus, the LLM may reproduce it instead of reasoning about alternatives.

By contrast, method-based reuse evaluates solutions based on logical applicability. Given a target question $Q_t$ with no direct solution, the framework identifies a candidate method $M_s = (Q_s, S_s)$ and applies $S_s$ to $Q_t$ if the following rationality condition holds:
\begin{equation}
    \text{Reuse}(Q_t, S_s) =
    \begin{cases}
        1 & \text{if } \text{Sim}_\text{logic}(Q_t, Q_s) \geq \tau, \\
        0 & \text{otherwise},
    \end{cases}
\end{equation}
where $\text{Sim}_\text{logic}$ measures logical similarity (e.g., general–specific or feature overlap) and $\tau$ is a threshold for rational applicability.

This ensures that reuse is not determined by token probability alone, but by whether the solution is logically transferable to the new context. For instance, although the statistical model may frequently associate ``color'' with fruit freshness, the logical reuse condition allows the method ``use picking time''—originally derived from a general fruit question—to be consistently reused for the specific banana question, as $Q_\text{banana} \subset Q_\text{fruit}$.

Formally, rational reuse can be expressed as:
\begin{equation}
    P_\text{reuse}(S_s \mid Q_t) \propto \text{Sim}_\text{logic}(Q_t, Q_s) \cdot \mathbb{I}[S_s \text{ valid}],
\end{equation}
where $\mathbb{I}[\cdot]$ is an indicator function ensuring that only logically valid methods are applied. This separates rational reuse from purely probabilistic word prediction, making the resulting solutions more interpretable and reliable.

\section{Verification} \label{sec_verification}
This section presents verification experiments for the proposed method reuse framework. The evaluation is divided into two parts: (1) relationship-based reuse, where questions exhibit clear vertical or horizontal relationships, and (2) feature-based reuse, where questions share only partial or hidden similarities without explicit structural relationships.

\subsection{Relationship Reuse}
The goal of this experiment is to verify whether methods can be effectively reused across questions that exhibit a clear general–specific (vertical) relationship. As a case study, we consider the question ``How to judge the freshness of a banana?'' and extend it using a more general method related to fruits. Here, fruit freshness is judged by \textit{picking time}, which differs from the methods directly suggested by ChatGPT-5. To avoid trivial recognition, the new method is embedded in surrounding text so that it is not directly highlighted.

We compare the proposed relationship-based method (\textbf{RelaMethod}) against a baseline method (\textbf{CompareRela}). The procedure for RelaMethod consists of:
(1) directly asking the LLM how to judge whether a banana is fresh,
(2) providing content that describes freshness evaluation for fruit based on picking time,
(3) prompting the LLM to separate the given content (in step (2)) into question–solution pairs, and
(4) asking the LLM again how to judge whether a banana is fresh.

In contrast, CompareRela follows steps (1), (2), and (4), but omits step (3), which is the proposed enhancement for exploiting vertical question–solution pairs. Both methods therefore receive the same information, but only RelaMethod explicitly structures it into reusable pairs.

\begin{figure}
    \centering
    \includegraphics[width=3.5in]{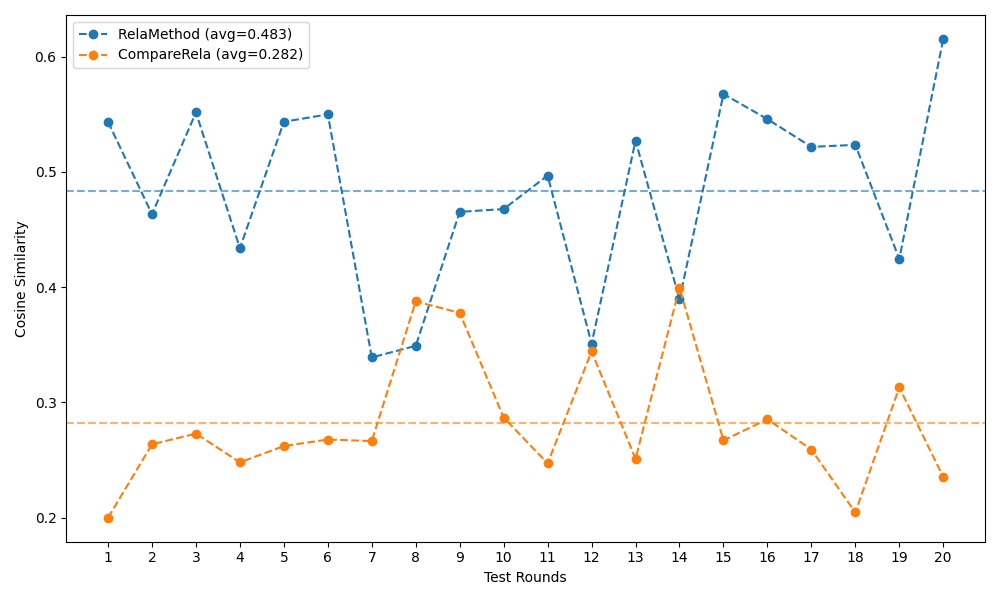}
    \caption{Cosine similarity comparison between RelaMethod and CompareRela.}
    \label{fig_relationship}
\end{figure}

Each method was tested over 20 rounds. Because LLM outputs may vary across runs, cosine similarity was used to measure alignment between the generated solutions and the target method (freshness judged by picking time). Only the relevant segments referring to picking time were compared to minimize unrelated noise. Results are shown in Fig.~\ref{fig_relationship}.

From Fig.~\ref{fig_relationship}, it can be observed that the relationship-based approach (RelaMethod), which explicitly prompts the LLM to generate question–solution pairs, consistently identifies the new method of using picking time to judge banana freshness. This is reflected in an average similarity score of 0.4835. In contrast, when the same materials are provided without explicit structuring (CompareRela), the LLM rarely recognizes picking time as a method, yielding a much lower average similarity of 0.2820.

The probability of recognizing the picking-time method in CompareRela is therefore low and largely inconsistent. For instance, in the 8th test round, CompareRela achieved a higher similarity than RelaMethod, but this occurred only because the LLM mentioned picking time as an ``extra note.'' Such incidental mentions appeared only 4 times across 20 trials, compared to 20 of 20 in RelaMethod, where picking time was consistently presented as an explicit method rather than as supplementary information.

To formally assess the difference, an independent-samples \textit{t}-test (without assuming equal variances) was conducted. RelaMethod achieved a mean similarity of 0.4835 with a standard deviation of 0.0801, whereas CompareRela achieved a mean of 0.2820 with a standard deviation of 0.0558. The analysis yielded a \textit{t}-value of 9.23 with a corresponding \textit{p}-value of $8.98 \times 10^{-11}$, which is far below the 0.05 threshold. These results confirm that RelaMethod consistently produces significantly higher similarity scores than CompareRela.

\subsection{Feature Reuse}
The objective of this verification is to evaluate whether methods can be reused across questions that do not exhibit explicit vertical or horizontal relationships, but instead share only partial or indirect similarities.

The test scenario is as follows. Tom wishes to reset the usage time of his hard disk by himself, without giving it to others, as he does not want anyone else to access its contents. However, he does not possess the necessary specialized tools. The question is therefore how he can accomplish this task on his own.

Although no identical cases exist, Tom has a prior experience (denoted as \textit{expMP3}) (without explicit vertical or horizontal relationships to the disk time reset question). His friend Mary once asked him to obtain a slower version of an MP3 file that was playing too fast. Tom was unable to download a slower version from several websites, and even after paying on the official MP3 site, he could only download a proprietary format rather than a standard MP3 file (likely due to copyright restrictions). Eventually, he solved the problem by purchasing the correct version from a goods-exchange website. By analogy, Tom later solved the hard disk problem in the same way—purchasing a disk usage time reset tool online—thus saving time and effort.

To prevent the expMP3 case from being too obvious, it was embedded among ten other stories of similar length and provided to the LLM as background material, following the procedure described in Section~\ref{sec_materials}.

Two methods were compared:
- \textbf{featureMethd}: (1) provide the 11 case studies; (2) ask the LLM, ``I need to reset my hard drive's usage time myself, but I lack the specialized tools to do so. How can I accomplish this without giving the drive to someone else?''; (3) prompt the LLM to generate question–solution pairs from the 11 case studies and then identify which, if any, could be applied to the hard disk reset issue.
- \textbf{compareMP3Method}: (1) provide the same 11 case studies; (2) ask the same disk-reset question; (3) prompt the LLM, ``Which of the 11 stories can give some indication for the disk time reset issues?’’—without explicitly requiring the generation of question–solution pairs.

\begin{figure}
    \centering
    \includegraphics[width=3.5in]{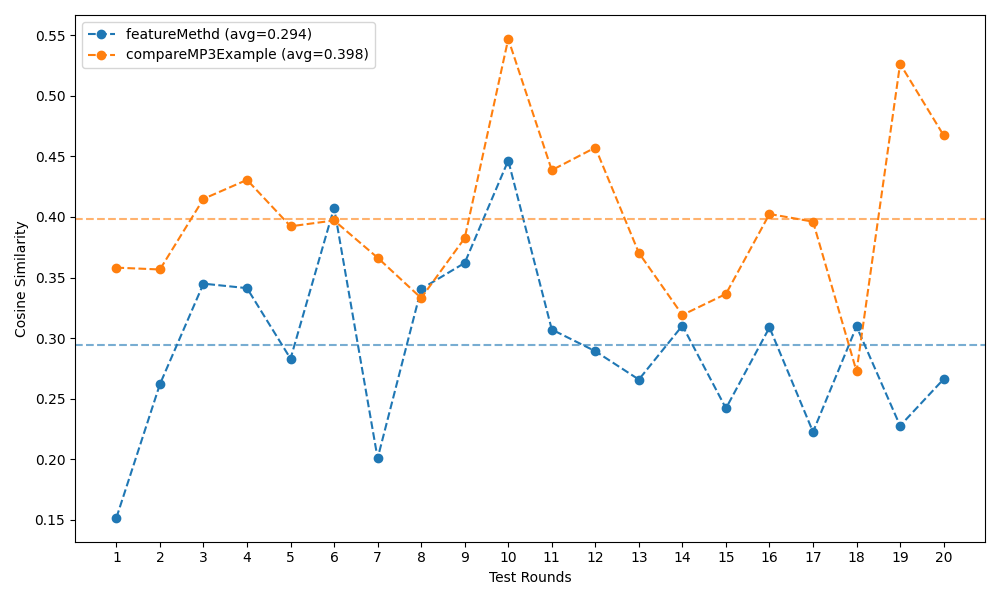}
    \caption{Cosine similarity comparison between featureMethd and compareMP3Method.}
    \label{fig_feature}
\end{figure}

Both methods were tested across 20 rounds, with results shown in Fig.~\ref{fig_feature}. Cosine similarity was again adopted as the evaluation metric, measuring whether the LLM reused the MP3-based method. Specifically, outputs were compared against the portion of text describing the purchase of a tool to resolve the MP3 issue, since this strategy was ultimately reused for solving the disk usage reset problem. By focusing on whether the LLM output contained this key method, we assessed the effectiveness of cross-question feature-based reuse.

From Fig.~\ref{fig_feature}, it can be observed that the feature-based approach (featureMethd), which explicitly prompts the LLM to separate questions and solutions, more frequently identifies the MP3-based strategy as reusable for the disk reset problem than the baseline method. This suggests that when the question and solution are explicitly decoupled, the LLM is more likely to transfer methods across cases that share only partial overlaps, even when the objects themselves do not belong to the same category. In this context, we use the term ``borrow'' to describe this type of relationship.

To quantify the difference between the two methods, a statistical analysis was conducted on the similarity scores. The featureMethd group achieved a mean similarity of 0.2945 with a standard deviation of 0.0698, while the compareMP3Method group achieved a higher mean of 0.3983 with a standard deviation of 0.0670. Welch’s \textit{t}-test was applied to evaluate the significance of this difference.
The analysis yielded a \textit{t}-value of $-4.80$ with a corresponding \textit{p}-value of $2.52 \times 10^{-5}$, which is far below the conventional threshold of 0.05. These results confirm that the two groups are statistically distinct, with compareMP3Method producing significantly higher similarity scores than featureMethd.

\subsubsection{Comparison of Relationship and Feature Reuse}
The experimental results reveal that relationship-based reuse achieves a clearer and stronger effect than feature-based reuse. Specifically, the mean difference between \textit{RelaMethod} and \textit{CompareRela} was 0.2015, nearly double the difference observed between \textit{featureMethd} and \textit{compareMP3Method} (0.1038). This finding is reinforced by the corresponding statistical tests: the RelaMethod–CompareRela comparison yielded a higher \textit{t}-value (9.23 vs.\ 4.80) and a lower \textit{p}-value ($8.98\times10^{-11}$ vs.\ $2.52\times10^{-5}$), indicating that the separation is both stronger and more reliable in the relationship-based setting.

As summarized in Table~\ref{tab:method_comparison}, this stronger distinction is also evident in relative ratios: 57.4\% for RelaMethod–CompareRela versus only 14.3\% for featureMethd–compareMP3Method. These results suggest that dependency-based reuse produces more stable and distinguishable similarity patterns, while feature-based reuse—relying only on partial correlations—remains comparatively weaker.

The underlying reason lies in the nature of similarity. The average similarity score for the RelaMethod–CompareRela pair was 0.3510, compared to just 0.0726 for the featureMethd–compareMP3Method pair (see Section~\ref{sec_sim_cal}). Relationship-based reuse exploits structural or dependency-driven connections, which lead to consistent recognition across test rounds. By contrast, feature-based reuse relies on partial overlaps that vary more across contexts. Consequently, the RelaMethod–CompareRela comparison exhibits a more pronounced and robust statistical separation.

\begin{table*}[htbp]
\centering
\caption{Comparison of Differences Between Method Pairs}
\begin{tabular}{lcccc}
\toprule
\textbf{Method Pair} & \textbf{Mean Difference} & \textbf{Average Similarity} & \textbf{Relative Ratio}  & \textbf{Reuse Type} \\
\midrule
\textit{RelaMethod} vs.\ \textit{CompareRela} & 0.2015 & 0.3510  & 57.4\% & Dependency-based reuse \\
\textit{featureMethd} vs.\ \textit{compareMP3Method} & 0.1038 & 0.0726  & 14.3\% & Partial correlation \\
\bottomrule
\end{tabular}
\label{tab:method_comparison}
\end{table*}

\section{Conclusion} \label{sec_conclusion}
This paper proposed a cross-question method reuse framework that extends the applicability of large language models (LLMs) beyond cases where questions are highly similar. By introducing relationship-based reuse, which leverages general–specific dependencies, and feature-based reuse, which captures partial or hidden similarities, the framework enables solutions to be transferred across a wider variety of problems. Verification experiments demonstrated that relationship-based reuse yields consistently higher and more statistically significant similarity scores compared to baselines, while feature-based reuse provides a complementary mechanism for handling less explicit overlaps. Together, these results highlight the potential of structured question–solution separation and reuse strategies to enhance the robustness and versatility of LLM-assisted problem solving.

\ifCLASSOPTIONcaptionsoff
  \newpage
\fi

\bibliographystyle{IEEEtran}
\bibliography{ref}

\section{Appendix} \label{sec_appendix}
\subsection{Experimental Materials (11 Case Studies)} \label{sec_materials}
The following 11 case studies were used as experimental materials for evaluating cross-question method reuse:

1. Lina found an old key buried in her grandmother’s garden. Nobody knew what it opened. One rainy afternoon, she discovered a hidden drawer in the attic desk. Inside lay yellowed letters and a photograph of her grandfather as a soldier. The key had preserved a secret love story, connecting Lina to her family’s forgotten past.

2. David missed his usual train home and caught the last one instead. The carriage was nearly empty except for an old man humming a lullaby. The melody was the same his mother used to sing. He sat quietly, listening. Somehow, the night felt less lonely, as if fate had guided him to that forgotten tune.

3. On a flooded street, a child placed a paper boat on the water. It floated past neighbors, carrying hope in the storm. Hours later, the boat reached an old man trapped upstairs, reminding him he wasn’t forgotten. Rescue soon arrived. He never knew the boy, but that fragile paper boat became his symbol of survival.

4. A friend had an MP3 file that was too fast, so she asked Tom to find a slower version for her. However, Tom couldn't download it from several websites as MP3  format. Even when he paid on the official MP3 site, he was unable to download the file. Finally, he tried to buy a copy from a goods exchange website.

5. Emma volunteered at the library every Saturday. One day, she noticed a man who never checked out books—he only sat, writing. Curious, she peeked at his notebook. Every page was filled with people’s faces he’d sketched, all drawn from memory. When she asked why, he whispered, “They’re people I loved. The library keeps them alive.”

6. A boy collected fireflies in a glass jar, pretending they were stars. His sick sister couldn’t see the night sky, so he brought the light to her room. She smiled, whispering, “You gave me the universe.” The fireflies flickered gently, as if agreeing. That small jar of light became their galaxy during the darkest nights.

7. Maya left her umbrella on a café chair. Hours later, a stranger picked it up. The next morning, he returned it with a note: “Umbrellas cover more than rain. Sometimes they cover loneliness.” Curious, Maya waited at the café again. The stranger returned, smiling. That umbrella became the beginning of conversations, and eventually, love.

8. Lucas found his grandfather’s broken pocket watch. It hadn’t ticked in years. He took it to a repair shop, where the watchmaker gently wound it. Suddenly, the watch sprang to life. Lucas smiled, feeling his grandfather’s presence. The ticking wasn’t just time—it was memory, echoing every story his grandfather had once told him.

9. On a quiet beach, Clara wrote “I forgive you” in the sand, hoping the waves would carry her message away. Hours later, she saw another phrase written nearby: “Thank you.” She never saw who wrote it, but the reply healed something inside her. Sometimes, forgiveness doesn’t need a face—only a whisper across the sea.

10. Every night, Mrs. Chen left her window lamp on. Her neighbor, a tired nurse, always passed by after work. That single light reminded her she wasn’t alone in the dark. One evening, the lamp didn’t glow. Worried, the nurse knocked on Mrs. Chen’s door, beginning a friendship. Sometimes, a small light sparks lasting bonds.

11. In an old attic, Sam discovered a dusty violin. Curious, he played a few notes, filling the silence with trembling sound. His grandmother climbed up, tears in her eyes. “That was your father’s. He stopped playing after the accident.” For the first time in years, music returned to their home—healing broken memories with melody.

\subsection{Materials Used for Feature Reuse Similarity} \label{sec_mat_simi_feat_reu}
'''Both involve blocked official tools.
   In the MP3 case, solution was find alternative but imperfect ways (other site, slower version).
   For disks: you cannot “reset” officially, but you can use alternatives (e.g., wipe data, replace drive, or use the disk differently).'''

\subsection{Materials Used for Relationship Reuse Similarity} \label{sec_sim_cal}
The following hints and questions were used to calculate similarity in the relationship and feature reuse experiments:

\textbf{Relationship reuse:}
\begin{itemize}
    \item Hint: ``The fruit picking time is the time when the fruit is picked, which can be recorded in blockchain to avoid alteration. It can also be used to judge whether a fruit is fresh. Fruits with shorter picking times are typically priced higher.’’
    \item Question: ``How to judge whether a banana is fresh or not?’’
\end{itemize}

\textbf{Feature reuse:}
\begin{itemize}
    \item Hint: ``A friend had an MP3 file that was too fast, so she asked Tom to find a slower version. However, Tom could not download it as MP3 format from several sites. Even after paying on the official site, only a proprietary format was available. Finally, he purchased a copy from a goods exchange website.’’
    \item Question: ``I need to reset my hard drive's usage time myself, but I lack the specialized tools to do so. How can I accomplish this without giving the drive to someone else?’’
\end{itemize}

%

\begin{IEEEbiography}{Hong Su}
  received the MS and PhD degrees, in 2006 and 2022, respectively, from Sichuan University, Chengdu, China. He is currently a researcher of Chengdu University of Information Technology Chengdu, China. His research interests include blockchain, cross-chain and smart contract.
\end{IEEEbiography}




\end{document}